\documentclass[11pt,a4paper]{article}
\usepackage[hyperref]{emnlp2018}
\usepackage{times}
\usepackage{latexsym}

\usepackage{url}

\usepackage{booktabs}
\usepackage{amssymb}
\usepackage[english]{babel}
\usepackage[utf8]{inputenc}
\usepackage{algorithm}
\usepackage[noend]{algpseudocode}
\usepackage{amsmath}
\usepackage{float}
\usepackage{graphicx}
\usepackage{caption}
\usepackage{subcaption}
\usepackage{enumitem}
\graphicspath{{figures/}}
\usepackage{relsize}

\aclfinalcopy 

\newcommand{\appendixhead}%
{\centering\textbf{\huge Appendix}
\vspace{0.25in}}

\begin{document}

\title{Dynamic Self-Attention: Computing Attention over Words Dynamically for Sentence Embedding}

\author{Deunsol Yoon\thanks{\, Equal Contribution. }\\
  Korea University \\
  Seoul, Republic of Korea \\
  {\tt emsthf930@naver.com} \\\And
  Dongbok Lee\footnotemark[1] \\
  Korea University \\
  Seoul, Republic of Korea \\
  {\tt markmarkhi@naver.com} \\\And
  SangKeun Lee \\
  Korea University \\
  Seoul, Republic of Korea \\
  {\tt yalphy@korea.ac.kr} \\}

\maketitle

\begin{abstract}

    In this paper, we propose Dynamic Self-Attention (DSA), a new self-attention mechanism for sentence embedding. We design DSA by modifying \textit{dynamic routing} in \textit{capsule network} \cite{capsule} for natural language processing. DSA attends to informative words with a dynamic weight vector. We achieve new state-of-the-art results among sentence encoding methods in Stanford Natural Language Inference (SNLI) dataset with the least number of parameters, while showing comparative results in Stanford Sentiment Treebank (SST) dataset.
    
\end{abstract}

\section{Introduction}\label{introduction}

In Natural Language Process (NLP), most neural network-based models contain a sentence encoder to map a sequence of words into a vector. The vector is then used for various downstream tasks, e.g., sentiment analysis, natural language inference, etc. The key part of a sentence encoder is a computation across a variable-length input sequence for a fixed size vector. One of the common approaches is the max-pooling in CNN or RNN \cite{yoonkim,infersent}.

Self-attention is another approach for a fixed size vector. Self-attention derived from the attention mechanism, originally designed for neural machine translation \cite{seqtoseqattention}, is utilized in various tasks \cite{Yang,hierarchical,varioustasks}. Self-attention computes attention weights by the inner product between words and the learnable weight vector. The weight vector is important in that it detects informative words, yet it is static during inference. The significance of the role of the weight vector casts doubt on whether its being static is an optimal status.


In parallel, \citet{capsule} recently proposed \textit{capsule network} for image classification. In \textit{capsule network}, \textit{dynamic routing} iteratively computes weights over inputs by the inner product between inputs and a weighted sum of inputs. Varying with the inputs, the weighted sum detects informative inputs; therefore it can be interpreted as a dynamic weight vector from the perspective of self-attention. We expect the dynamic weight vector to give rise to flexibility in self-attention since it can adapt to given sentences even after training.



Motivated by \textit{dynamic routing} \cite{capsule}, we propose a new self-attention mechanism for sentence embedding, namely Dynamic Self-Attention (DSA). To this end, we modify \textit{dynamic routing} such that it functions as self-attention with the dynamic weight vector. DSA, which is stacked on CNN with Dense Connection \cite{dense}, achieves new state-of-the-art results among the sentence encoding methods in Stanford Natural Language Inference (SNLI) dataset with the least number of parameters, while obtaining comparative results in Stanford Sentiment Treebank (SST) dataset. It also outperforms recent models in terms of time efficiency due to its simplicity and highly parallelized computations.

\begin{figure*}
        \centering
        \includegraphics[width=\textwidth]{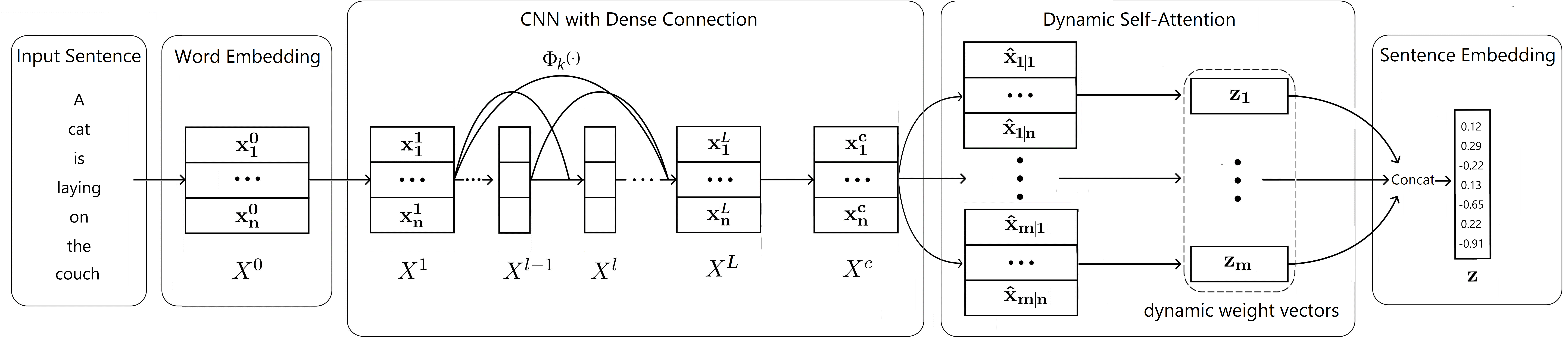}
        \caption{Overall architecture of our approach. ${\bf z_1, ..., z_m}$ are dynamic weight vectors. The final sentence embedding, i.e. $\bf z$, is generated by concatenating ${\bf z_1, ..., z_m}$.}\label{fig:model}
\end{figure*}

Our technical contributions are as follows:
\noindent\begin{itemize}[parsep=0pt,labelsep=*,leftmargin=1pc]
\item We design and implement Dynamic Self-Attention (DSA), a new self-attention mechanism for sentence embedding.
\item We devise the dynamic weight vector with which DSA computes attention weights.
\item We achieve new state-of-the-art results in SNLI dataset, while showing comparative results in SST dataset. 

\end{itemize}

\section{Preliminary}\label{back}
In self-attention \cite{Yang,hierarchical}, attention weights are computed as follows:
\begin{equation}
{\bf a} = \text{Softmax}({\bf v^T}\text{Tanh}(WX))\label{eq1}
\end{equation}
where $X\in \mathbb{R}^{d_w \times n}$ is an input sequence, $W\in \mathbb{R}^{d_v \times d_w}$ is a projection matrix and ${\bf v}\in \mathbb{R}^{d_v}$ is the learnable weight vector of self-attention. The weight vector ${\bf v}$ plays an important role, since attention weights are computed by the inner product between $\bf v$ and the projection of the input sequence $X$. The weight vector ${\bf v}$ is static with respect to the input sequence $X$ during inference. Replacing the weight vector $\bf v$ with a weight matrix enables multiple attentions \cite{self-attentive,disan}.  

\section{Our Approach}\label{ourapproach}
Our architecture, shown in Figure~\ref{fig:model}, is built on CNN with Dense Connection \cite{dense}. Dynamic Self-Attention (DSA), which is stacked on CNN with Dense Connection, computes attention weights over words.

\subsection{CNN with Dense Connection}\label{module1}
The goal of this module is to encode each word into a meaningful representation space while capturing local information. We do not add any positional encoding, as suggested by \citet{convseq2seq}; deep convolution layers capture relative position information. We also enforce every output of layers to have the same number of columns by using appropriate zero padding.

We denote a sequence of word embeddings as ${X^{0}}\in \mathbb{R}^{d_0 \times n}$, where ${X^{0}} = [\bf {x^{0}_{1}}, {\bf x^{0}_{2}}, ... , {\bf x^{0}_{n}}]$. $h^l(\cdot)$ is a composite function of the $l^{th}$ layer, composed of 1D Convolution, dropout \cite{dropout}, and leaky rectified
linear unit (Leaky ReLU). We feed a sequence of word embeddings into $h^1(\cdot)$ with kernel size 1:
\begin{equation}
{X}^{1} = {h^1}(X^0)\label{eq2}
\end{equation}
where ${X^{1}}\in \mathbb{R}^{d_{1} \times n}$. We add Dense Connection in every layer $h^l(\cdot)$ with the same kernel size:
\begin{align}
{X}^{l} = {h^{l}}(concat[X^{l-1},X^{l-2},...,X^{1}])\label{eq3}
\end{align}
where ${X^{l}}\in \mathbb{R}^{d_{l} \times n}$, and $l\in[2,L]$. We concatenate outputs of all $h^l(\cdot)$, and denote it as a single function:
\begin{equation}
\Phi_k(X^0) = concat[X^L, X^{L-1}, ..., X^1]\label{eq4}
\end{equation}
where kernel sizes of all $h^l(\cdot)$ in $\Phi_k(\cdot)$ are the same number $k$, except for $h^1(\cdot)$. We then feed outputs of two different functions $\Phi_{k_1}(\cdot), \Phi_{k_2}(\cdot)$, and a sequence of word embeddings $X^0$ into a compression layer:
\begin{equation}
{X}^{c} = h^{c}(concat[\Phi_{k_1}(X^0), \Phi_{k_2}(X^0), X^0])\label{eq5}
\end{equation}
where $h^{c}(\cdot)$ is the composite function with kernel size 1. It compresses the first dimension of input (i.e., ${2\sum_{l=1}^{L} d_{l} + d_0}$) into $d_{c}$ to represent a word compactly. Finally, $L_2$ norm of every column vector $\bf{x^{c}_i}$ in the $X^c$ is normalized, which is found to help our model to converge fast and stably.

\subsection{Dynamic Self-Attention (DSA)}\label{module2}
Dynamic Self-Attention (DSA) iteratively computes attention weights over words with the dynamic weight vector, which varies with inputs. DSA enables multiple attentions in parallel by multiplying different projection matrices to ${X^{c}}$, the output from CNN with Dense Connection. For the $j^{th}$ attention, DSA projects the compact representation of every word $\bf x^{c}_i$ with LeakyReLU activation:
\begin{equation}
{\bf {\hat x}_{j|i}} = \text{LeakyReLU}(W_j {\bf x^{c}_i} + {\bf b_j})\label{eq6}
\end{equation}
where ${W_j}\in \mathbb{R}^{d_{o} \times d_{c}}$ is a projection matrix, ${\bf b_j} \in \mathbb{R}^{d_o}$ is a bias term for the $j^{th}$ attention. Given the number of attentions $m$, i.e., $j\in[1,m]$, attention weights of words are computed by following Algorithm~\ref{alg:euclid}:

\begin{algorithm}[H]
\centering
\caption{Dynamic Self-Attention}\label{alg:euclid}
\begin{algorithmic}[1]
\Procedure{Attention}{${\bf {\hat x}_{j|i}}, r$}
\State for all $i^{th}$ word, $j^{th}$ attention : $q_{ij} = 0$
\For{r iterations}
\State for all $i,j$ : ${a_{ij} = \frac{exp(q_{ij})}{{\sum_k}{exp(q_{kj})}}}$
\State for all $j$ : ${\bf s_{j}} = \sum_i a_{ij}{\bf {\hat x}_{j|i}}$
\State for all $j$ : ${\bf z_{j}} = \text{Tanh}({\bf s_{j}})$\label{query}
\State for all $i,j$ : $q_{ij} = q_{ij} + {\bf{\hat x}_{j|i}^T}{\bf z_{j}}$
\EndFor\label{euclidendwhile}
\State \textbf{return} all $\bf z_j$
\EndProcedure
\end{algorithmic}
\end{algorithm}

$r$ is the number of iterations, and $a_{ij}$ is the attention weight for the $i^{th}$ word in the $j^{th}$ attention. $\bf{z_j}$ is the output for the $j^{th}$ attention of DSA at the $r^{th}$ iteration, and also the dynamic weight vector for the $j^{th}$ attention of DSA before $r^{th}$ iteration. The final sentence embedding ${\bf z}$ is the concatenation of $\bf{z_1}, ..., \bf{z_m}$:
\begin{equation}
{\bf z} = concat[\bf{z_1}, ..., \bf{z_m}]\label{eq7}
\end{equation}
where ${\bf z}\in \mathbb{R}^{md_{o}}$ is used for downstream tasks.

We modify \textit{dynamic routing} \cite{capsule} to make it function as self-attention with the dynamic weight vector. We remove capsulization layer in \textit{capsule network} which transforms scalar neurons to capsules, multi-dimensional neurons. A single word is then not decomposed into multiple capsules, but represented as a single vector ${\bf x^{c}_i}$ in Eq.~\ref{eq6}. \textit{Squashing} function is a nonlinear function for capsules. We replace it with Tanh nonlinear function for scalar neurons in Line~\ref{query} of Algorithm~\ref{alg:euclid}. We also force all the words in the $j^{th}$ attention to share a projection matrix $W_{j}$ in Eq.~\ref{eq6}, as an input is a variable-length sequence. By contrast, each capsule in \textit{capsule network} has its own projection matrix $W_{ij}$.

\subsection{Dynamic Weight Vectors}\label{model3}

\begin{figure}
        \centering    
        \includegraphics[width=0.5\textwidth]{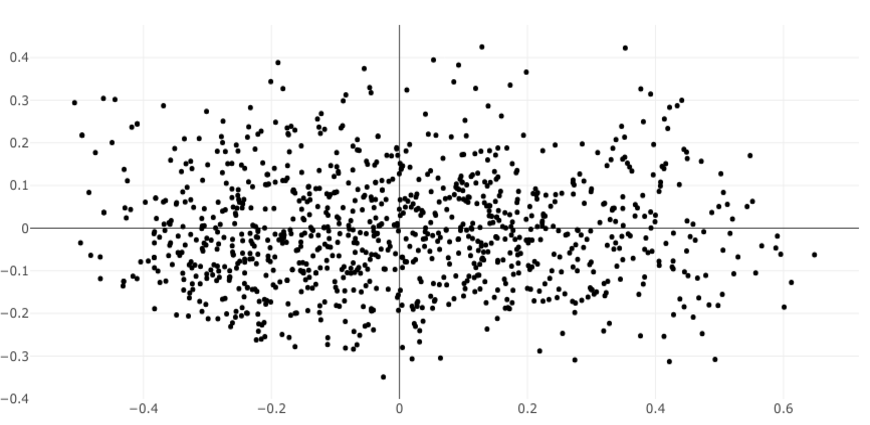}
        \caption{Dynamic weight vectors visualization}\label{fig:pca}
\end{figure}

The weight vector ${\bf v}$ of self-attention in Eq.~\ref{eq1} is static during inference. In DSA, however, the dynamic weight vector $\bf{z_j}$ in Line~\ref{query} of Algorithm~\ref{alg:euclid}, varies with an input sequence $\bf {\hat x}_{j|1}, ..., {\hat x}_{j|n}$, even after training. In order to show how the dynamic weight vectors vary, we perform dimensionality reduction on them, $\bf z_{1}$ at the $(r-1)^{th}$ iteration of Algorithm~\ref{alg:euclid}, by Principal Component Analysis (PCA). We randomly select 1,000 sentences from Stanford Natural Language Inference (SNLI) dataset and plot each dynamic weight vector for the sentences in 2D vector space. Figure~\ref{fig:pca} shows that dynamic weight vectors are scattered in all directions. Thus, DSA adapts the dynamic weight vector with respect to each sentence. 


\section {Experiments}\label{experiments}

We evaluate our sentence embedding method with two different tasks: natural language inference and sentiment analysis. We implement single DSA, multiple DSA and self-attention in Eq.~\ref{eq1} as a baseline. Both DSA and self-attention are stacked on CNN with Dense Connection for fair comparison.

For our implementations, we initialize word embeddings by 300D GloVe 840B pretrained vectors \cite{glove}, and fix them during training. We use cross-entropy loss as an objective function for both tasks. We set $d_o=600$, $m=1$ for single DSA and $d_o=300$, $m=8$ for multiple DSA. In Appendix, we provide details for training our implementations, hyperparameter settings, and visualization of attention maps of DSA.



\setlength{\tabcolsep}{0.4em}
\begin{table*}[t]
	\small
	\begin{center}
		\begin{tabular}{|l|c|c|c|c|}
		
			\hline
			{\bf Model}                & {\bf Train (\%)} & {\bf Test (\%)}  & {\bf Parameters (m)} & {\bf T(s)/epoch}  \\ 
			\hline
			600D BiLSTM with self-attention \cite{Yang}                & 84.5    & 84.2 & 2.8  & -      \\
			300D Directional self-attention network \cite{disan}                 & 91.1    & 85.6 & 2.4  & { 587} \\
			600D Gumbel TreeLSTM~\cite{gumble}                 & 93.1    & 86.0 & 10.0  &  -    \\
			600D Residual stacked encoders \cite{residual}              & 91.0    & 86.0 & 29.0  &    -   \\
			300D Reinforced self-attention network \cite{reinforce}                & 92.6    & 86.3 & 3.1  & 622\\
			1200D Distance-based self-attention network \cite{distance}               & 89.6    & 86.3 & 4.7  & 693  \\
		    \hline
		    600D CNN (Dense) with self-attention & 88.7 & 84.6 & {2.4} &  121     \\
		    ours (600D Single DSA)   & 87.3 & 86.8 & \textbf{2.1}  & 135      \\
		    ours (2400D Multiple DSA)   & 89.0    & \textbf{87.4} & {7.0}  & 198      \\
		    \hline
		\end{tabular}
	\end{center}
	\caption{SNLI Results. The values in T(s)/epoch come from original papers and are experimented on the same graphic card to ours (single Nvidia GTX 1080Ti). Word embedding is not counted in parameters.}
	\label{tab:SNLI}
\end{table*}



\subsection{Natural Language Inference Results}


Natural language inference is a task of classifying the semantic relationship between two sentences, i.e., a premise and a hypothesis. We conduct experiments on Stanford Natural Language Inference (SNLI) dataset, consisting of human-written 570k pairs of English sentences labeled with one of three classes: Entailment, Contradiction and Neutral. As the task considers the semantic relationship, SNLI is used as a benchmark for evaluating the performance of a sentence encoder.

We follow a conventional approach, called \textit{heuristic matching} \cite{heuristicmatching}, to classify the relationship of two sentences. The sentences are encoded by our proposed model. Given encoded sentences ${\bf s^h}, {\bf s^p}$ for hypothesis and premise respectively, an input of the classifier is $concat[{\bf s^h}, {\bf s^p}, |{\bf s^h} - {\bf s^p}|, {\bf s^h}\odot{}{\bf s^p}]$.


The results from the official SNLI leader  board\footnote{\url{https://nlp.stanford.edu/projects/snli/}} are summarized in Table~\ref{tab:SNLI}. Single DSA achieves new state-of-the-art results with test accuracy (86.8\%) and the number of parameters (2.1m). Besides, our learning time per epoch (135s) is significantly faster than recent models because of its simple structure and highly parallelized computations. With tradeoffs in terms of parameters and learning time per epoch, multiple DSA outperforms other models by a large margin (+1.1\%).

In comparison to the baseline, single DSA shows better performance than self-attention (+2.2\%). This confirms that the dynamic weight vector is more effective for sentence embedding. Note that our implementation of the baseline, self-attention stacked on CNN with Dense Connection, shows better performance (+0.4\%) than the one stacked on BiLSTM \cite{Yang}.

\setlength{\tabcolsep}{0.4em}
\begin{table}[t]
	\small
	\begin{center}
		\begin{tabular}{|l|c|c|}
		    \hline
			{\bf Model} & {\bf SST-2} & {\bf SST-5} \\ 
			\hline
			BiLSTM \cite{baseline_lstm_bilstm} & 87.5  & 49.5  \\
			CNN-non-static \cite{yoonkim} & 87.2 & 48.0\\
			BiLSTM with self-attention  & 88.2   & 50.4 \\
		    CNN (Dense) with self-attention  & 88.3  & \bf{50.6} \\
			\hline
		    ours (Single DSA) & \textbf{88.5} & \textbf{50.6} \\
		    \hline
		\end{tabular}
	\end{center}
	\caption{Test accuracy with SST dataset.}
	\label{tab:SST}
\end{table}

\subsection {Sentiment Analysis Results}
Sentiment analysis is a task of classifying sentiment in sentences. We use Stanford Sentiment Treebank (SST) dataset, consisting of 10k English sentences, to evaluate our model in single-sentence classification. We experiment SST-2 and SST-5 dataset labeled with binary sentiment labels and five fine-grained labels, respectively. 

The SST results are summarized in Table~\ref{tab:SST}. We compare single DSA with four baseline models: BiLSTM \cite{baseline_lstm_bilstm}, CNN \cite{yoonkim} and self-attention with BiLSTM or CNN with dense connection. Single DSA outperforms all the baseline models in SST-2 dataset, and achieves comparative results in SST-5, which again verifies the effectiveness of the dynamic weight vector. In contrast to the distinguished results in SNLI dataset (+2.2\%), in SST dataset, only marginal differences in the performance between DSA and the previous self-attentive models are found. We conclude that DSA exhibits a more significant improvement for large and complex datasets.


\section{Related Works}\label{relatedworks}


Our work differs from early self-attention for sentence embedding \cite{Yang,hierarchical,self-attentive,disan} in that the dynamic weight vector is not static. Independently, there have recently been an approach to \textit{capsule network}-based NLP. \citet{capsuletext2} applied whole \textit{capsule network} to text classification task. However, we only utilize an algorithm, \textit{dynamic routing} from \textit{capsule network}, and modify it into self-attention with the dynamic weight vector, without unnecessary concepts, e.g., capsule.

\section{Conclusion}\label{conclusion}

In this paper, we have proposed Dynamic Self-Attention (DSA), which computes attention weights over words with the dynamic weight vector. With the dynamic weight vector, the self attention mechanism can be furnished with flexibility. Our experiments show that DSA achieves new state-of-the-art results in SNLI dataset, while showing comparative results in SST dataset.

\bibliography{dsa.bbl}
\bibliographystyle{acl_natbib_nourl}

\twocolumn[\appendixhead]
\section*{Detailed Experimental Settings}\label{supplemn}
\addcontentsline{toc}{section}{Appendices}
\renewcommand{\thesubsection}{\Alph{subsection}}

For training our implementations in both Standford Natural Language Inference (SNLI) dataset and Stanford Sentiment Treebank (SST) dataset, we initialize word embeddings by 300D GloVe 840B pretrained vectors, and fix them during training. We use cross-entropy loss as an objective function for both tasks. For both tasks, we set $r=2$, $k_1=3$, $k_2=5$, $L=4$, $d_1=150$, $d_l=75$, $\text{where } l\in[2,L]$, and $d_c=300$. All models are implemented via PyTorch. Details of the hyperparameters for each task are introduced in each section. Note that we followed data preprocessing of SST as \cite{yoonkim}\footnote{\url{https://github.com/yoonkim/CNN_sentence}} and SNLI as \cite{infersent}\footnote{{\url{https://github.com/facebookresearch/InferSent}}}

\subsection{SNLI}

We minimize cross-entropy loss with Adam optimizer. We apply $m=1,d_o=600$ for single DSA and $m=8,d_o=300$ for multiple DSA. We use two-hidden layer multilayer perceptron (MLP) with leaky rectified
linear unit (Leaky ReLU) activation as a classifier, where the number of hidden layer neurons are 300 (single) or 512 (multiple). Batch normalization and dropout are used for an input for a hidden layer in the MLP. Dropout rate is set to 0.3 (single) or 0.4 (multiple) in the MLP. For the both models, we use 0.00001 L2 regularization. We use dropout with rate of 0.2 in every $h^{l}(\cdot)$, and $h^{c}(\cdot)$. We also use dropout with rate of 0.3 for word embeddings. We initialize parameters in every layer with He initialization  and multiply them by square root of the dropout rate of its layer. We initialize out-of-vocabulary words with $\mathcal{U}(-0.005, 0.005)$. starting with default value of Adam, we halve learning rate if training loss is not reduced for five times with a patience of 0.001. The size of mini-batch is set to 256.

\begin{figure*}
        \centering
        \includegraphics[width=\textwidth]{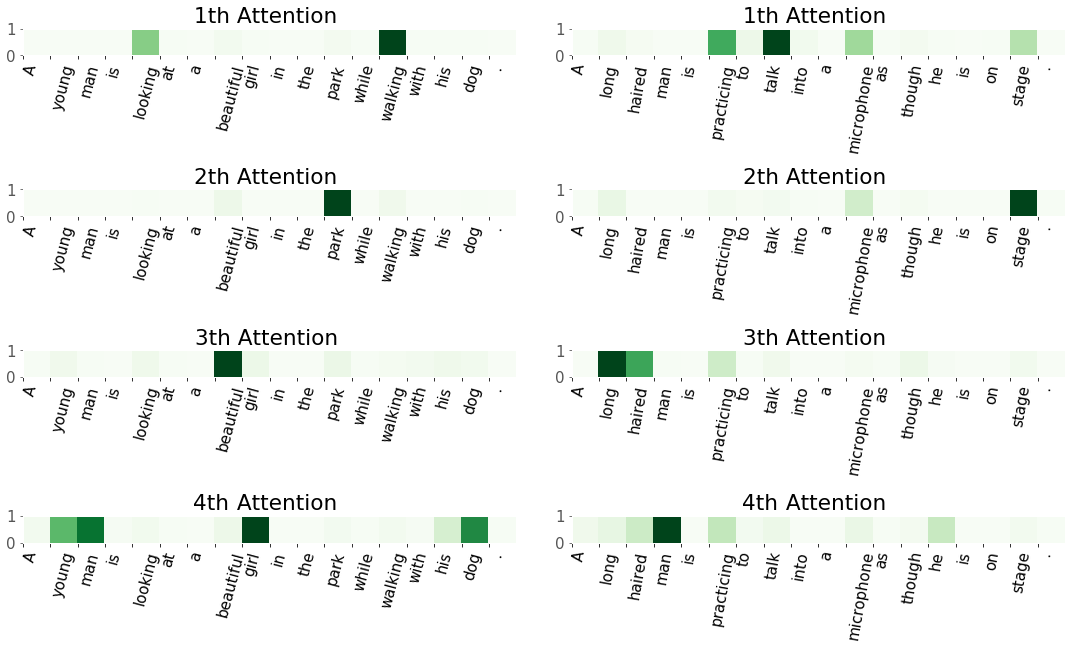}
        \caption{We visualize 4 out of 8 attentions from multiple DSA, which are human interpretable. Each attention in multiple DSA attends different aspects in the given sentences. $1^{th}$ attention only attends words related to a verb, $2^{th}$ attends related to a place, $3^{th}$ attends related to adjective, and $4^{th}$ attends related to an organism.}
\end{figure*}

\subsection{SST}
We minimize cross-entropy loss with Adadelta optimzier. We apply $d_o=600, m=1$. We use one-hidden layer MLP with Leaky ReLU activation as a classifier, where the number of hidden layer neurons is 300. Batch normalization and dropout with rate of 0.4 are used for an input for a hidden layer in MLP. For regularization, we use 0.00001 L2 regularization. We use dropout with rate of 0.2 in every $h^{l}(\cdot)$ and $h^{c}(\cdot)$. We also use dropout with rate of 0.4 for word embeddings. We initialize parameters in every layer with He initialization and multiply them by square root of the dropout rate of each layer. We initialize out-of-vocabulary words with $\mathcal{U}(-0.05, 0.05)$. Starting with default value of Adadelta, we halve learning rate halved learning rate if training loss is not reduced for two times with a patience of 0.001. The size of mini-batch is set to 128.

\section*{Visualization}

\begin{figure}[H]
        \includegraphics[width=0.5\textwidth]{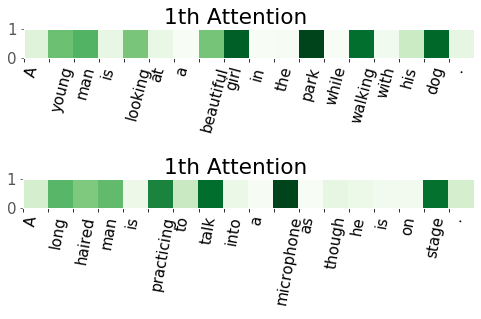}
        \caption{Single DSA attends only informative words in the given sentences.}
\end{figure}

\end{document}